\documentclass{article}

     \PassOptionsToPackage{numbers, compress}{natbib}

    \usepackage[preprint]{neurips_2024}

\usepackage[utf8]{inputenc} 
\usepackage[T1]{fontenc}    
\usepackage{hyperref}       
\usepackage{url}            
\usepackage{booktabs}       
\usepackage{amsfonts}       
\usepackage{amsmath}
\usepackage{amsthm}
\usepackage{nicefrac}       
\usepackage{microtype}      
\usepackage{xcolor}         
\usepackage{graphicx}
\usepackage{float}
\usepackage{subcaption}

\newtheorem{prn}{Principle}
\title{Time Matters: Scaling Laws for Any Budget}

\author{
  Itay Inbar\\
  Google DeepMind \\
  \texttt{inbaritay@google.com} \\
  \And
  Luke Sernau \\
  Google DeepMind\\
  \texttt{sernau@google.com} \\
}

\begin{document}

\maketitle

\begin{abstract}
A primary cost driver for training large models is wall-clock training time. We show that popular time estimates based on FLOPs are poor estimates, and construct a more accurate proxy based on memory copies. This allows us to accurately estimate the training speed of a transformer model from its hyperparameters. Combined with a scaling law curve like Chinchilla, this allows us to accurately predict the final loss of a model from a simple equation. We show that this expression is accurate across a wide range of model hyperparameter values, enabling us to analytically make architectural decisions and train models more efficiently. Crucially, this analysis predicts that in contrast to existing literature, models should be wider rather than deeper, as the benefits of speed outweigh the benefits of depth.
\end{abstract}

\section{Introduction}
The final quality of a language model is constrained by the number of parameters and the amount of data it was trained on. Remarkably, these two parameters alone are often sufficient to estimate the final performance of the model. \citet{kaplan2020scaling} explored this phenomenon, predicting that loss curves during pretraining could be written as a linear combination of a term dependent on the number of the parameters and one dependent on the dataset size. \citet{hoffmann2022training} refined this estimate, improving the estimation of the coefficients and introducing a bias term to capture the inherent perplexity of language.

While these estimates are useful for large-scale models, small and mid-sized models are not at risk of running out of pretraining data. Instead, the limiting factor is the cost of training, a figure which is primarily driven by a model's size and speed. This suggests that instead of trading off model size and dataset size, we should be trading off architectural hyperparameters within the model that affect its throughput. On a fixed budget, a faster model will be able to see more tokens than a slow one.

In this work we assume a fixed training time, and ask what hyperparameters we should pick to maximize the final performance of the model. We start by estimating the throughput of the model (tokens per second) in terms of the number of FLOPs and memory copies, both of which can be directly calculated from the model's hyperparameters.

\citet{kaplan2020scaling} mentions a parameterization in terms of compute requirements, but their estimate is based on FLOPs, which we will show are a weak predictor of runtime. Instead, we show that memory copies are a much stronger predictor. This predictor, while simplistic, is powerful enough to accurately predict the loss in terms of the hyperparameters of the model.

This new framing lets us estimate the final loss of a model without training it, given only the model hyperparameters and the desired training time. We show that this method produces accurate predictions across a wide range of hyperparameter values, and makes useful predictions about which hyperparameters should be used in order to maximize training efficiency.

We evaluate our findings over 1,535 different decoder-only transformer models configurations ranging from $300K$ to $310M$ parameters and trained over the C4 dataset \cite{dodge2021documenting}. We achieve an $r^2$ of 0.9 when predicting their final loss using our refined scaling law. This is the same $r^2$ we get when using the traditional Chinchilla scaling law. In other words, we are able to estimate the final loss with the same accuracy whether we use Chinchilla scaling laws on empirical runtimes or simply estimate them from hyperparameters.

\section{The parameter equivalence heuristic}
The core intuition motivating this work is the observation that large models are not particularly sensitive to their hyperparameters, provided we hold the total parameter count constant. This idea was discussed in \cite{kaplan2020scaling}, but due to its importance we capture it in a form of an equivalence heuristic.

\begin{prn}[The Parameter Equivalence Heuristic]
Above a certain scale, the final loss of a transformer at the end of training is primarily a function of how many parameters there are, not where they are in the model.
\end{prn}

One straightforward implication is that we ought to be able to predict the final loss using only parameter count and number of training tokens, as earlier scaling laws did. But another often overlooked implication is that models of the same size that allocate their parameters differently compete primarily on speed.

If it is not feasible for one model to have vastly lower loss than another via architectural improvements, we should instead choose architectures that optimize for training speed, allowing them to consume as many tokens during training time as it can. We can show this in practice by means of a scaling law.

\section{Estimating linear scaling law Coefficients}\label{scaling_coeff_sct}
The original scaling law from \citet{hoffmann2022training}, predicts the final training loss of a language model in terms of its parameters count \(N\) and the number of tokens it was trained upon \(D\).
\begin{equation}\label{orig_chinchila}
L(N,D) = \frac{A}{N^\alpha} + \frac{B}{D^\beta} + E
\end{equation}
\citet{hoffmann2022training} and \citet{besiroglu2024chinchilla} derive different coefficient values with the most extreme difference being their linear data coefficient \(B\). Rather than enter into this debate, we simply take the exponents from \cite{hoffmann2022training}, and fit our own linear coefficients \(A\), \(B\) and \(E\) using linear regression on the model loss. This was done by iterating over 767 different decoder-only transformer models' hyperparmeters configurations trained from scratch for three hours each on the C4 \citet{dodge2021documenting} dataset and then evaluating our predictions over a holdout set of 767 different model configurations trained in the same manner over the same dataset. Models sizes vary from $319K$ to $310M$ parameters. Note that we constrained our experiments to models that can be trained on a single TPU to avoid confounders from inter-chip communication. We experimented with model hyperparameters of embed sizes ranging from $2^5$ to $2^{10}$, number of layers ranging from $3$ to $8$, MLP width ranging from $2^8$ to $2^{14}$, number of attention heads ranging from $2^1$ to $2^7$, and a fixed vocabulary size of 8,000. We trained on a mesh of 4 hosts x 8 chips/host of TPU V5 chips with no model sharding.

\begin{figure}[H]
  \centering
  \includegraphics[width=.65\linewidth]{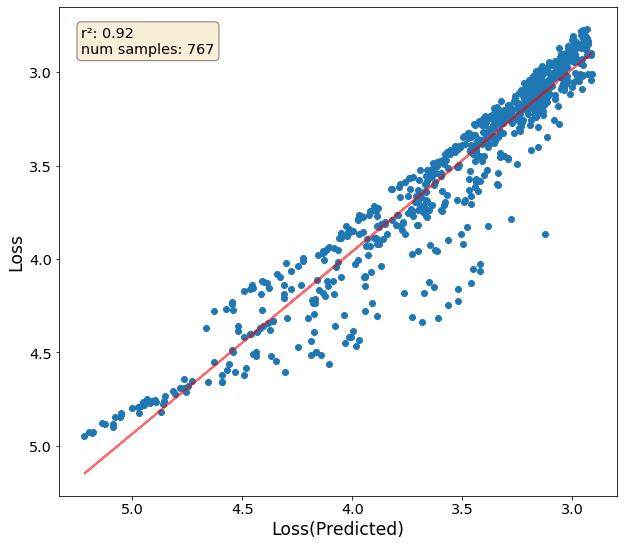}
  \caption{Loss predictions over different trained models via Chinchilla, using our linear coefficients.}
  \label{ABE_loss_fig}
\end{figure}

The results show a very good fit (\(r^2=0.9\)), with coefficients \(A=195.76\), \(B=182.52\), and \(E=2.34\). Note that these are different from the values quoted in either paper \cite{hoffmann2022training}\cite{besiroglu2024chinchilla}.

Using the values from the papers presents a very different story. The below table compares the scaling law fitting measurements on our data using the different papers coefficients, with our computed coefficients serving as a baseline.

\begin{table}[H]\label{slopes_tbl}
\caption{Comparing scaling coefficients} 
\begin{center} 
\begin{tabular}{ |c|c|c|c| } \hline
 \textbf{Fitted slope} & \textbf{Fitted intercept} & $\mathbf{r^2}$ & \textbf{Source} \\ \hline 
 0.48 & 1.50 & 0.9 & Hoffmann et al. \\  \hline 
 0.43 & 1.70 & 0.9 & Besiroglu et al. \\  \hline  
 0.48 & 1.57 & 0.9 & Besiroglu et al. table params \\  \hline  
 1.00 & 0.00 & 0.9 & Hoffmann et al. with our A,B,E \\  \hline  
\end{tabular} 
\end{center} 
\end{table} 

Both Chinchilla papers underestimate the loss by a factor of more than two on our data (\(\text{slope}<0.5\)). We take this as evidence that these coefficients are perhaps highly sensitive to the details of the setup, possibly explaining the discrepancy between the papers. Nonetheless, we were able to achieve very good fit with a linear rescaling of their predictions (\(r^2=0.9\) in all cases), suggesting the exponents are more robust.

\section{Equations for estimating the speed of a model}
In order to use scaling laws to estimate the loss of a model, we need to know how big the model is and how much data will it be able to train over. The former is a straightforward exercise in accounting, but the latter is more nuanced.

We fix the amount of time we have to train the model to some constant $T$, which is measured in seconds. If we can estimate how long each training step takes for a given model, we can work out how many tokens (data) will be processed by that model in time \(T\).

It is tempting to imagine that we could estimate the model training speed just by adding up the number of FLOPS. But as we will show, the runtime of the model is actually driven by data copying, not the actual computation.

The amount of data copying depends on a wide variety of factors, from the hardware to the architecture to the compiler. We do not attempt to account for all of these factors here but take as a simplifying assumption that every matrix multiplication requires a copy proportional to the size of its operands.

Specifically, for a standard (as defined in \cite{vaswani2017attention}) decoder-only transformer architecture, we derived equations for the number of parameters a model has (PARAMS), the number of memory loads the model will need to make in a single pass (MEMCPYS), and the number of operations the model will do in a single pass (FLOPS):

\begin{align*}
    &\text{PARAMS}\left(d,n,v,w\right) = vd + nd\left(8 + 2w + 4d\right) + nw\\
    &\text{MEMCPYS}\left(d,n,s,v,w\right) = 2vd + 2sv + ns\left(w + 2hs \right) + 2nd\left(w + 4s +                                         2d\right)\\
    &\text{FLOPS}\left(d,n,s,v,w\right) = 2svd + 2dns\left(w + 2d + s\right) + nhs^2
\end{align*}

Where the parameters are defined as:

$\qquad d = \text{embedding dimension}$

$\qquad n = \text{number of layers}$

$\qquad s = \text{sequence length}$

$\qquad v = \text{vocabulary size}$

$\qquad w = \text{MLP width}$

$\qquad h = \text{number of heads}$

The full details of the derivation of the above equations can be found in appendix \ref{appendix_equations}.

Using a linear combination of the above equations we can now compute the total number of seconds per training step (TIME) as:

\begin{equation}\label{time}
\text{TIME}\left(d,n,s,v,w\right) = c_1\text{MEMCPYS}\left(d,n,s,v,w\right) + c_2\text{FLOPS}\left(d,n,s,v,w\right) + c_3
\end{equation}
Where \(c_1\), \(c_2\), and \(c_3\) are coefficients determined by linear regression (see Section \ref{runtime}).
Note that dividing the total number of seconds per training step (i.e., TIME) by the number of seconds we are training upon (i.e., $T$) would yield the total number of training steps (i.e., D in \eqref{orig_chinchila}).

Finally, we can estimate the total loss by plugging the above term into the Chinchilla scaling law in order to derive an equation dependent on model training speed.
\begin{equation}\label{loss}
\hat{L}\left(d,n,s,v,w\right) = E + \frac{A}{\text{PARAMS}\left(d,n,v,w\right)^\alpha} + B\left(\frac{\text{TIME}\left(d,n,s,v,w\right)}{T}\right)^\beta
\end{equation}

Following our findings in Section \ref{scaling_coeff_sct}, we take the original \(\alpha\) and \(\beta\) as in \cite{hoffmann2022training} and use our own fitted linear coefficients for \(A\), \(B\) and \(E\).

\section{Estimating the throughput}\label{runtime}
The throughput of a model is defined to be \(1/\text{TIME}\), where \(\text{TIME}\) is defined in Equation \eqref{time}. In order to fully specify equation \eqref{time} we need to determine its linear coefficients \(c_1\), \(c_2\), and \(c_3\).

We conduct a larger scale (\(N=1,778\)) sweep over model hyperparameters trained for 5 minutes, just long enough to accurately determine the number of tokens per second they process. We applied linear regression over a holdout set of the same size(\(N=1,778\)) to determine that \(c_1=3.74e-19\), \(c_2=2.4e-15\), and \(c_3=1.46e-07\).

We trained models of sizes varying from $277K$ parameters to $972M$ parameters. We experimented with model hyperparameters of embed sizes ranging from $2^5$ to $2^{12}$, number of layers ranging from $1$ to $8$, MLP width ranging from $2^8$ to $2^{15}$, number of heads ranging from $2^0$ to $2^7$, and a fixed vocabulary size of 8,000. As in previous experiments we trained on a mesh of 4x8 TPU V5 chips with no model sharding.

\begin{figure}[H]
  \centering
  \includegraphics[width=.65\linewidth]{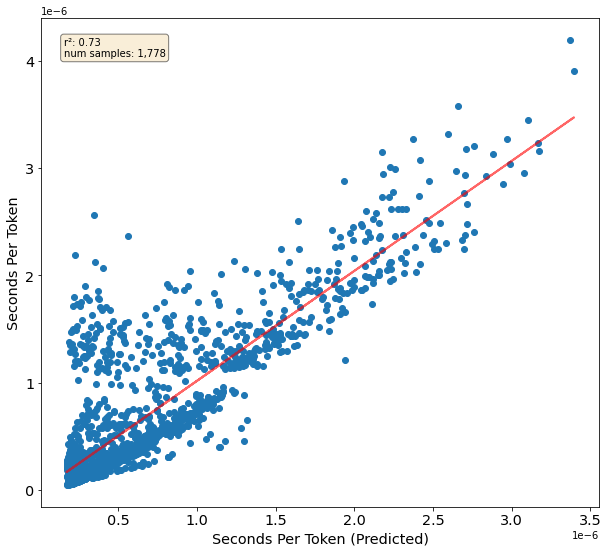}
  \caption{Estimating runtime with equation \eqref{time}}
  \label{runtime_prediction_fig}
\end{figure}

The results show an overall \(r^2\) of \(0.74\), with a much tighter fit for slower (i.e. bigger) models. For fast models, confounding factors like compiler optimizations start to matter, affecting the quality of the fit.

It is worth evaluating the importance of the different terms(i.e. FLOPS, MEMCPY) in Equation \eqref{time}. Previous work by both \cite{hoffmann2022training} and \cite{besiroglu2024chinchilla} utilized only the FLOPS counting to derive their scaling laws. We show that MEMCPY is a stronger predictor, and can account for essentially all of the explanatory power on its own.

\begin{figure}[H]
  \centering
  \includegraphics[width=.75\linewidth]{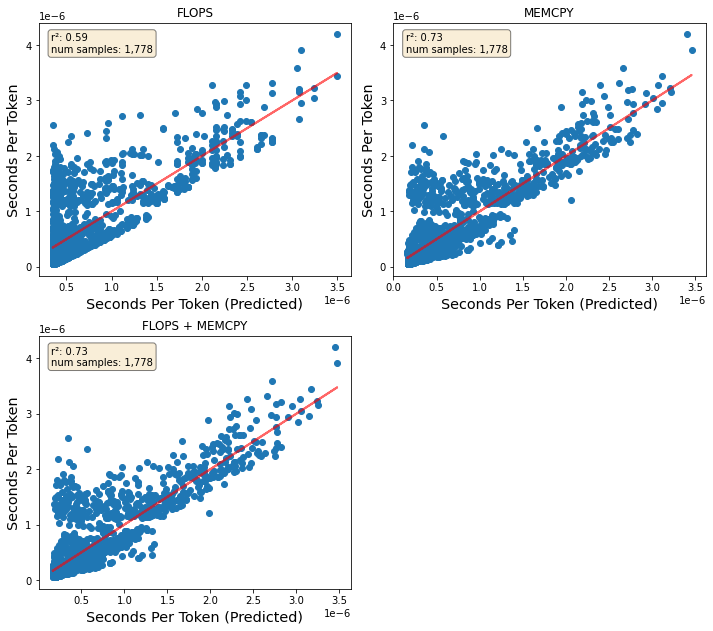}
  \caption{Runtime prediction ablation}
  \label{throughput_ablation}
\end{figure}

\section{Putting it all together}
\begin{figure}[H]
  \centering
  \includegraphics[width=.5\linewidth]{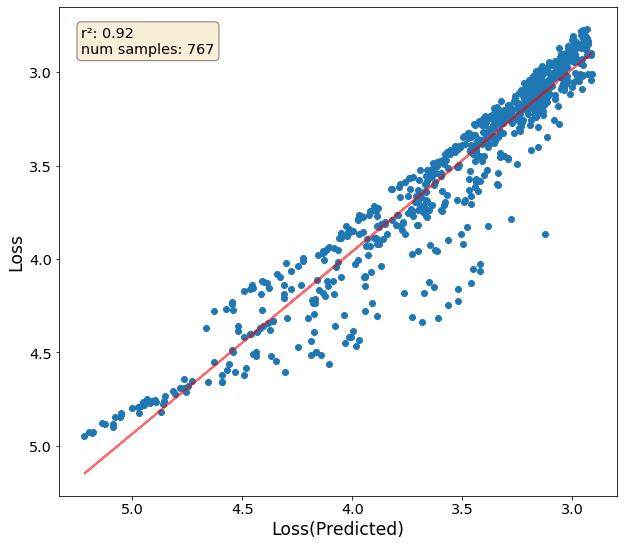}
  \caption{Predicted vs actual loss. The prediction is made using only the hyperparameters.}
  \label{L_N_T_fig}
\end{figure}

We now have an equation that estimates the number of tokens that the model will consume from its hyperparameters. We also have an exact expression for the number of parameters in such a model, PARAMS. Our tuned Chinchilla equation relates these two quantities to estimate the final loss \eqref{loss}. In figure \ref{L_N_T_fig}, we show the results of this estimation, applied to the holdout data from Section \ref{scaling_coeff_sct}.

Notice that the graph is largely indistinguishable from Figure \ref{ABE_loss_fig}, including the quality of fit \(r^2 = 0.92\). While there is some error in the Chinchilla equation's predictions, there is essentially no additional error from using our estimates in place of the empirical values.

\begin{figure}[H]
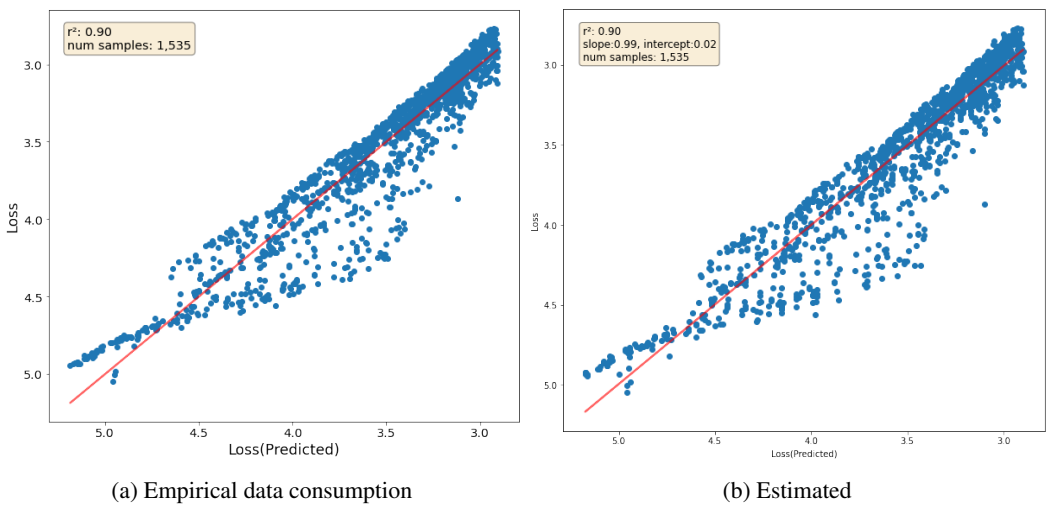

\begin{subfigure}{0.5\textwidth}
  \centering
  \includegraphics[width=\linewidth]{ABE_loss_flip.png}
  \caption{Empirical data consumption}
\end{subfigure}%
\begin{subfigure}{0.5\textwidth}
  \centering
  \includegraphics[width=\linewidth]{L_N_T_flip.png}
  \caption{Estimated}
\end{subfigure}
\caption{Chinchilla using empirical data consumption vs estimated (ours)}
\label{hidd_dim_fig}
\end{figure}

\section{Better loss with faster models}
\begin{figure}[H]
  \centering
  \includegraphics[width=0.85\linewidth]{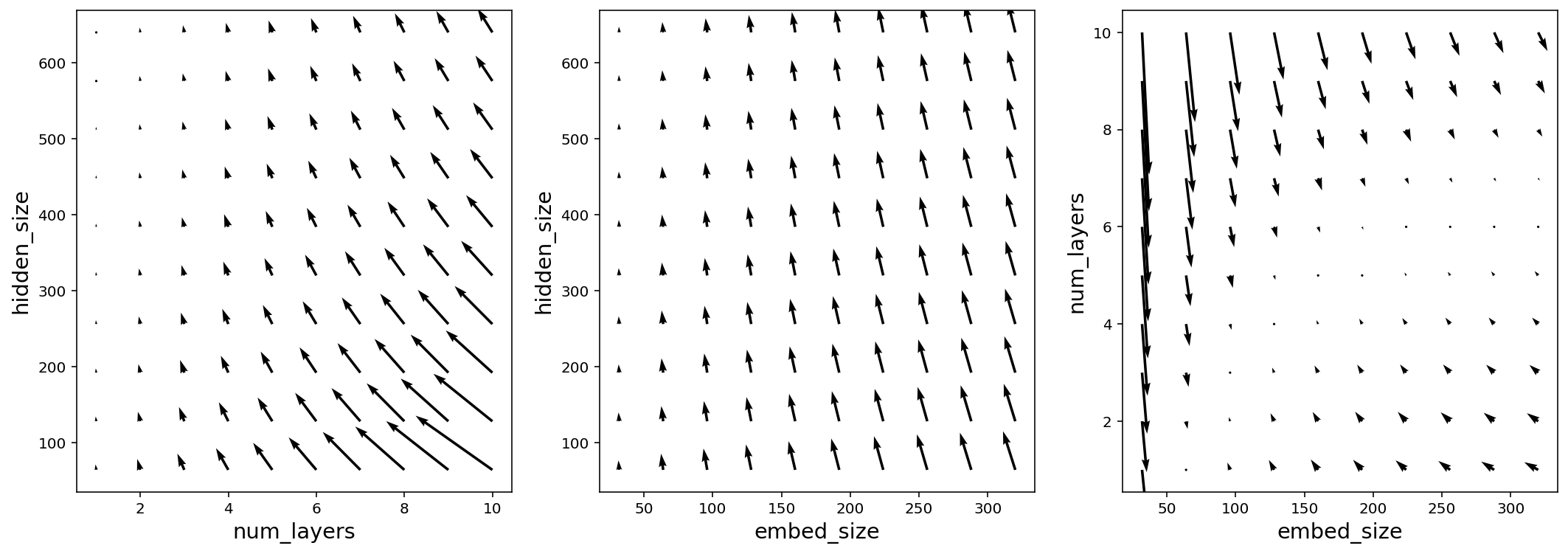}
  \caption{Negative gradient of the loss, projected such that the arrows point in directions of constant parameter count. That is, following an arrow corresponds to moving in the direction of steepest decent subject to the constraint that the parameter count is held constant.}
  \label{gradients_fig}
\end{figure}

We can use these equations to make specific predictions about how we should size our models. Figure \ref{gradients_fig} shows the negative gradient of the loss with respect to each of our hyperparameters, projected to be along level curves of the parameter count. Following each arrow brings you to another model with the same parameter count but a lower predicted loss.

We can see that increasing the hidden size at the expense of the other hyperparameters is favorable throughout the plotted region. \texttt{num\_layers} is particularly disincentivized. This suggests that in contrast with common practice we should take our MLPs to be wide, and our models to be somewhat shallow, in exchange for smaller embed size.

\section{Other Architectures}
There are many architectures other than transformers for which this kind of technique could be of interest. We focused on transformers as they are the dominant architecture used in practice today. With that said, our core observations are simply that
\begin{itemize}
    \item  If there exists a scaling law that can predict loss in terms of the number of parameters and the amount of data, the lowest loss will be achieved when we arrange the parameters to make the model as fast as possible, allowing us to train on more data.
    \item Estimating speed correctly requires considering not only the FLOPs but also the MEMCPYs.
    \item FLOPs, MEMCPYs and the total number of parameters can all be counted using simple accounting.
\end{itemize}
These observations are not specific to transformers, even if the particulars of the accounting differs. For this reason, we expect analogous results for other architectures.

\section{Conclusion}
Understanding what hyperparameters lead to the strongest model performance is a vital part of model design. We've shown that the final loss of a model can be accurately predicted by turning the question on its head. Instead of asking for the most data efficient hyperparameters, we simply ask which hyperparameters make the model the fastest. This leads to a new scaling law based on hyperparameters alone. In the long run, the faster model will tend to win.

We demonstrated this effect across a wide variety of model sizes, and showed that we can accurately predict the model's loss from its hyperparameters, simply by estimating how many memory copies will take place. Crucially, this is a stronger predictor than approaches based on FLOPs. In particular, it predicts that in contrast with common practice we should be making shorter, wider models, because the speed benefits dominate in practice.

However, we do not consider the effects of model sharding, or the effects of scale beyond a few hundred million parameters. We regard these as fruitful areas of exploration for future work.

\bibliography{TimeMatters}


\appendix

\section{Equations}\label{appendix_equations}
\subsection{FLOPS derivation}
In order to compute the total number of FLOPS in our transformers decoders stack we begin by counting the FLOPS needed for each step in a transformer block.

\begin{table}[H]\label{t1}
\caption{Transformer FLOPS} 
\begin{center} 
\begin{tabular}{ |c|c| }  
\hline 
 $3sd^2$ & Produce $Q$, $K$ and $V$ inside the attention \\  \hline 
 $h\left(s^2d/h\right)$ & Compute $QK^T$ \\  \hline  
 $h\left(s^2\right)$ & Apply softmax \\  \hline  
 $h\left(s^2  d/h\right)$ & Multiply by $V$ \\  \hline  
 $sd^2$ & Projection layer to recombine the heads \\  \hline  
 $sdw + swd$ & The two layers of the MLP \\  \hline  
\end{tabular} 
\end{center} 
\end{table} 

We sum all of these components and multiply by the number of transformer layers in our transformer stack.

The final piece of the puzzle is to add the embedding of the input and the output. Both of which require $svd$ FLOPS. We add all of these terms together and simplify. 

\subsection{MEMCPYS derivation}
We begin by adding up the total amount of data being copied in a single transformer block. We approximate the number of memory copies needed for each matmul as the size of the input matrices for each operation.

\begin{table}[H]\label{m1}
\caption{MEMCPYS} 
\begin{center} 
\begin{tabular}{ |c|c| }  
\hline 

 $3\left(sd + d^2\right)$ & Produce $Q$, $K$ and $V$ inside the attention \\  \hline 
 $h\left(sd/h + sd/h\right)$ & Compute $QK^T$ \\  \hline  
 $h\left(s^2\right)$ & Apply softmax \\  \hline  
 $h\left(s^2 + sd/h\right)$ & Multiply by $V$ \\  \hline  
 $sd+ d^2$ & Projection layer to recombine the heads \\  \hline  
 $sd + dw + sw + wd$ & MLP \\  \hline  
\end{tabular} 
\end{center} 
\end{table} 

Again we sum all of the above components and multiply by the number of transformer layers in our transformer stack.

Finally, we add the embedding of the input and the output, both of which require $v*d + s*v$ memory copies. Summing all of this together and simplifying yields our MEMCPY equation.

\subsection{PARAMS derivation}
We begin with the per-layer parameters. We note the extra vector term to account for the bias term accompanying each matrix.

\begin{table}[H]\label{p1}
\caption{PARAMS} 
\begin{center} 
\begin{tabular}{ |c|c| }  
\hline 
 $3\left(d + d^2\right)$ & $Q$, $K$ and $V$ inside the attention \\  \hline 
 $4d$ & Layer norms before and after the attention \\  \hline  
 $d^2 + d$ & Projection layer to recombine the heads \\  \hline  
 $2dw + d + w$ & MLP matrices \\  \hline
\end{tabular} 
\end{center} 
\end{table} 

In similar fashion, we left with summing all of the above components and multiplying by the number of transformer layers in our transformer stack.\\
The final piece of the puzzle is to add a $2d$ vector for the norm layer after the transformers as well as the embedding matrix used for embedding the input and the output, matrix of size $vd$. Summing all of these and simplifying yields our PARAMS equation.

\end{document}